\documentclass[english]{lni}

\IfFileExists{latin1.sty}{\usepackage{latin1}}{\usepackage{isolatin1}}

\usepackage{graphicx}
\usepackage{fancyhdr}
\usepackage{listings} 
\usepackage{changepage} 
\usepackage[figurename=Fig., tablename=Tab.]{caption}[2008/04/01]

\author{\c{S}amil Karahan\textsuperscript{1}\,Merve K{\i}l{\i}n\c{c} Y{\i}ld{\i}r{\i}m\textsuperscript{1}\,Kadir K{\i}rta\c{c}\textsuperscript{1}\,Ferhat \c{S}\"{u}kr\"{u} Rende\textsuperscript{1}\\ \,G\"{u}ltekin B\"{u}t\"{u}n\footnote{TUBITAK BILGEM, Gebze, Kocaeli, TURKEY, \{samil.karahan, merve.kilinc, kadir.kirtac, ferhat.rende, gultekin.butun\}@tubitak.gov.tr} \,Haz{\i}m Kemal Ekenel\footnote{Faculty of Computer and Informatics, Dept. of Computer Engineering, 34469 Maslak, Istanbul, TURKEY, ekenel@itu.edu.tr} 
}

\title{How Image Degradations Affect Deep CNN-based Face Recognition?}
\begin{document}
\maketitle

\renewcommand{\refname}{References}
\setcounter{footnote}{2} 

\begin{abstract}
Face recognition approaches that are based on deep convolutional neural networks (CNN) have been dominating the field. The performance improvements they have provided in the so called in-the-wild datasets are significant, however, their performance under image quality degradations have not been assessed, yet. This is particularly important, since in real-world face recognition applications, images may contain various kinds of degradations due to motion blur, noise, compression artifacts, color distortions, and occlusion. In this work, we have addressed this problem and analyzed the influence of these image degradations on the performance of deep CNN-based face recognition approaches using the standard LFW closed-set identification protocol. We have evaluated three popular deep CNN models, namely, the AlexNet, VGG-Face, and GoogLeNet. Results have indicated that blur, noise, and occlusion cause a significant decrease in performance, while deep CNN models are found to be robust to distortions, such as color distortions and change in color balance. 
\end{abstract}

\section{Introduction}
Due to its paramount role in application areas, such as biometrics and surveillance, face recognition has become a very popular research area in computer vision community. With the recent advances in deep convolutional neural networks, researchers have reached encouraging improvements in face recognition accuracy on the LFW \cite{huang2007labeled} and YouTube Faces \cite{wolf2011face} datasets \cite{sun2015deepid3,parkhi2015deep,schroff2015facenet}. However, a systematic analysis is required to assess the performance of deep CNN-based approaches under different appearance variations in order to find out their strengths and weaknesses with respect to these variations and point out future research directions according to the outcomes of these experiments. In the literature, the effects of quality factors such as contrast, brightness, sharpness, focus, and illumination on commercial face recognition system \cite{dutta2012impact} and shallow representation based face recognition \cite{abaza2014design} are studied. The experiments for both studies show that the effects of quality deformation on recognition performance is significant. Besides face recognition, image quality effects are also studied in different classification applications, such as object classification \cite{dodge2016understanding} and hand-written digit recognition \cite{basu2015learning}.

In this paper, we present a systematic evaluation to assess the effect of image quality, which is listed as one of the main factors that limits the performance of the face recognition algorithms \cite{jain2011handbook}, on deep CNN-based face recognition approaches' performance.
\begin{figure}[htb]
\centering
\IfFileExists{SampleGaussianEffect-crop.pdf}{}{\immediate\write18{pdfcrop SampleGaussianEffect.pdf}}
\includegraphics[width=\textwidth,keepaspectratio]{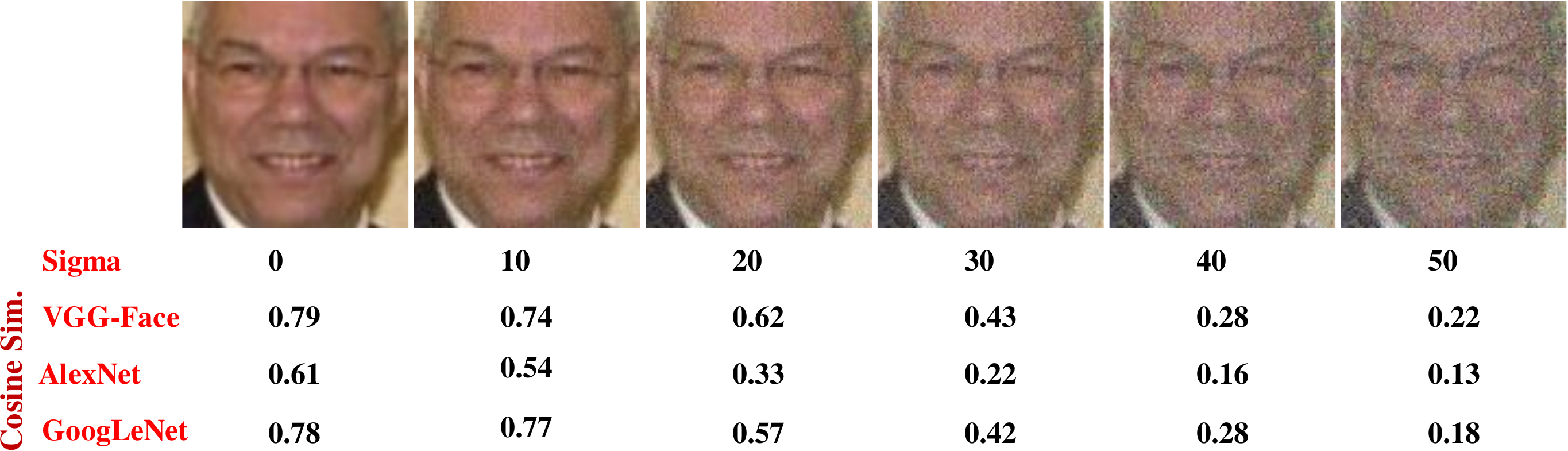}
\caption{A sample face image distorted by adding Gaussian noise. Cosine similarity values were obtained by comparing features of the distorted sample and a fixed gallery sample of the same subject, extracted from three deep CNN-based models. Similarity value of 1.0 is perfect match}
\label{fig:SampleGaussianEffect}
\end{figure}
We have applied image degradations based on contrast, noise, blur, occlusion, and color distortions at different levels to the probe set of LFW closed set identification protocol, whereas gallery images have been kept in their original form. We perform a thorough analysis of effect of these quality degradations on the state-of-the-art deep CNN-based face representations. Three well-known, successful deep convolutional neural network models, VGG-Face \cite{parkhi2015deep}, GoogLeNet \cite{szegedy2015going}, and AlexNet \cite{krizhevsky2012imagenet} have been used. We also point that further research is required to robustify deep CNN models based face recognition against image blur, noise, and occlusion. 

\pagestyle{fancy}
\fancyhead{} 
\fancyhead[RO]{\small How Image Degradations Affect Deep CNN-based Face Recognition? \hspace{5pt} \thepage \hspace{0.05cm}}
\fancyhead[LE]{\hspace{0.05cm}\small \thepage \hspace{5pt} \c{S}amil Karahan et al.}
\fancyfoot{} 
\renewcommand{\headrulewidth}{0.4pt} 

In Fig.\ref{fig:SampleGaussianEffect}, we present cosine similarity values of a sample face image distorted by adding Gaussian noise with different standard deviations. Similarity values were obtained by comparing features of the distorted sample and a fixed gallery sample of the same subject, extracted from three popular deep CNN-based models. The similarity value of 1.0 means perfect match. It is clear from the figure that when distortion level (sigma value) increases, cosine similarity values for all the networks decrease significantly.

The main contributions of this work can be listed as follows: (i) We provide a standard benchmark to assess face recognition algorithms' performance under various image degradations. (ii) We have extensively evaluated and compared performance of CNN-based face recognition approaches' under these variations. We show that our fine-tuned GoogLeNet model is able to achieve state-of-the-art closed-set identification results. Fine-tuned GoogLeNet is found to be superior to fine-tuned AlexNet and VGG-Face models. (iii) We have pointed that image blur, noise, and occlusion pose significant challenges to the deep CNN-based face recognition models. These models are found to be robust to image degradations due to color distortions and change in color balance. Overall, we believe that this study will bring useful insights to develop face recognition algorithms that are robust to visual distortions that can occur in real-world environments.

\section{Deep Convolutional Neural Network Architectures}
Three different, commonly used deep CNN architectures have been utilized in this work. These are the AlexNet \cite{krizhevsky2012imagenet}, VGG-Face \cite{parkhi2015deep}, and GoogLeNet \cite{szegedy2015going}. AlexNet and GoogLeNet have been trained on ImageNet \cite{krizhevsky2012imagenet}, which contains 1.2M training images of 1,000 object categories. AlexNet and GoogLeNet models have been fine-tuned by using pre-trained models that were released by Caffe library \cite{jia2014caffe}. Fine-tuning has been done using a combination of the following datasets: CASIA-Webface \cite{yi2014learning}, FaceScrub \cite{ng2014data}, and CACD \cite{chen2014cross}. Dataset preparation contains manual effort for correcting mislabeling and duplicate removal. The final dataset contains about 530K images of 10,770 unique subjects.

The parameters for fine-tuning AlexNet were as follows: \%15 of the dataset is randomly selected as the validation set, while the rest is used as the training set. Optimization is done via Stochastic Gradient Descent (SGD) with momentum using the Caffe library \cite{jia2014caffe}. Momentum is set to 0.9, and weight decay is set to 5e-4 in our experiments. Learning rate is set to 1e-3 initially and decreased by a scaling factor of 0.1 in every 20,000 iterations. Training batch size is set to 128. Maximum number of iterations is set to 100,000 to ensure that number of training epochs will be 30. Fine-tuning was performed on a standard PC with NVIDIA TITAN X GPU. We have used the 4096-dimensional feature vector extracted from the \textit{fc7} layer of the AlexNet in our experiments.

The parameters for fine-tuning GoogLeNet were as follows: Learning rate is set to 1e-3 initially and decreased by a scaling factor of 0.15 in every 84,000 iterations. Fine-tuning was performed on a machine with two NVIDIA K20 GPUs (a total of 10GB GPU memory), using a maximum allowed training batch size of 32. Maximum number of iterations is set to 420,000 to ensure that the number of training epochs will be 30. The rest of the settings are the same with AlexNet settings. We use the 1024-dimensional feature vector extracted from the \textit{pool5/7x7\_s1} layer of the GoogLeNet in our experiments.

We have employed the 4096-dimensional feature vector extracted from the \textit{fc6} layer of the VGG-Face network [PVZ15] in our experiments.

\section{Experimental Setup}
We have applied the LFW closed-set identification protocol as our face recognition benchmark \cite{best2014unconstrained}. In this protocol, the gallery set contains 4249 subjects, with a single image per subject, and the probe set contains 3143 face images of the same subjects in the gallery. We have used cosine distance as similarity metric \cite{taigman2015web}. Distortion effects were only applied to the probe set. We have considered five popular distortion types; blur, noise, contrast, color distortion, and synthetic occlusion generated on fixed facial parts. 

\begin{figure}[htb]
\IfFileExists{SampleImages-crop.pdf}{}{\immediate\write18{pdfcrop SampleImages.pdf}}
\includegraphics[width=\textwidth,keepaspectratio]{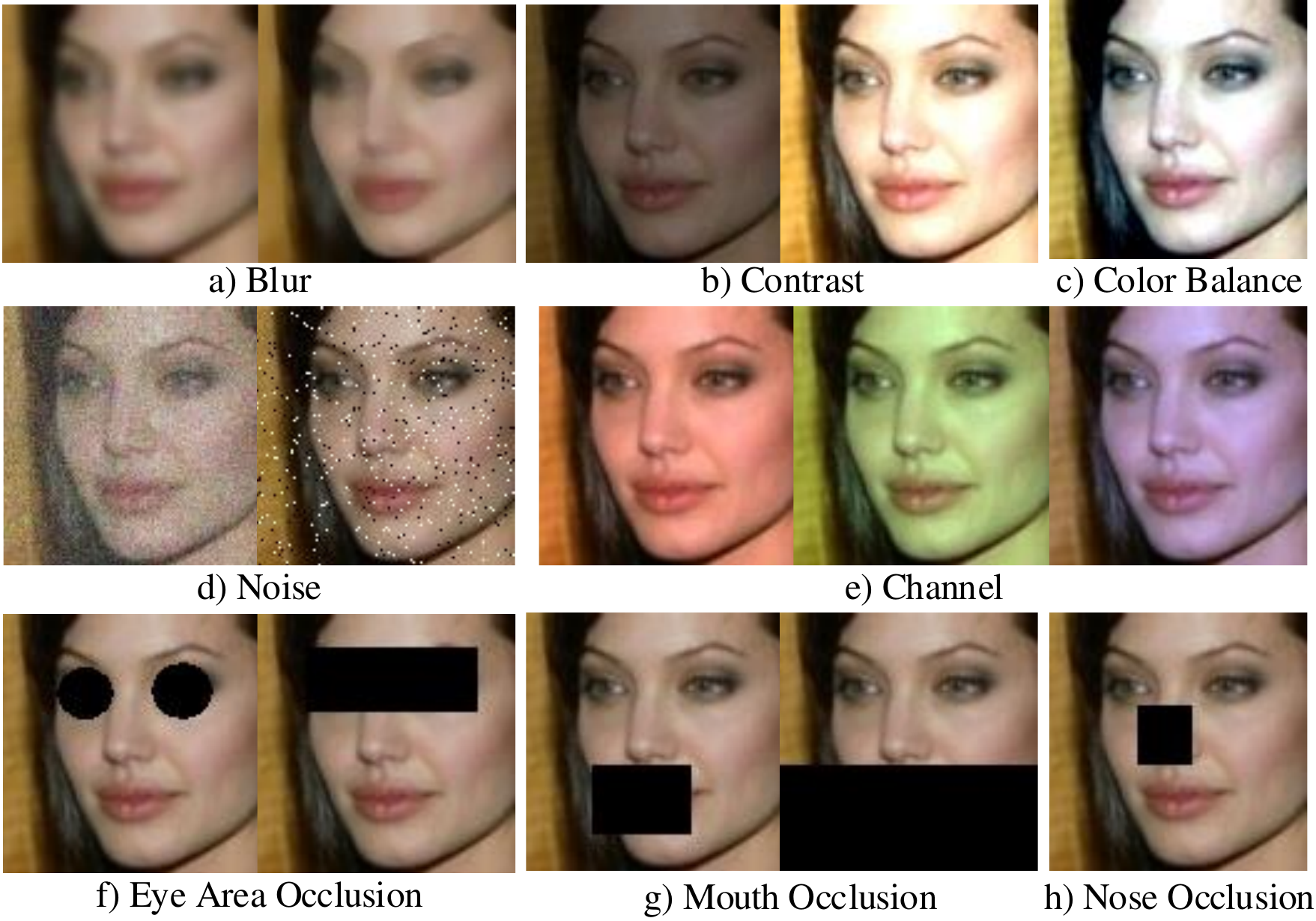}
\caption{a) left: Gaussian blur (sigma = 2.0, kernel size = 7), right: median blur (kernel size = 7), b) left: low contrast, right: high contrast, c) the color balance of the original is changed at the rate of 20 percent, d) left: Gaussian noise (sigma = 30), right: salt and pepper noise (param=6), e) the intensity of each channel is increased 60 pixel, f) left: the sun glasses effect, right: darkened periocular region, g) left: mouth occlusion, right: the scarf effect, h) nose area is occluded.}
\label{fig:SampleImages}
\end{figure}

To simulate blur effect, we have applied Gaussian filter and median filter functions. The sigma value of the Gaussian filter is varied between one and five in steps of one. The kernel size is adjusted with respect to the sigma, which is four times minus one of the sigma value. The kernel size of median filter is varied from three to nine in steps of two (see Fig.\ref{fig:SampleImages}-a).

To simulate the contrast effect, we multiply each channel value with constant values varied from 0.2 to 2.0 in steps of 0.2. In this work, we aim to experiment both the decrease and increase in contrast (see Fig.\ref{fig:SampleImages}-b).

We have added noise to images by using the two most popular noise models; Gaussian and Salt and Pepper noise. To add Gaussian noise, we have first created normally distributed random numbers with zero mean and varying standard deviation from 10 to 50 in steps of 10. The obtained result is normalized to [0-255] intensity range. To simulate the Salt and Pepper noise, for each pixel, we first generate a uniformly distributed random number between 0 and 255. We then repeat the following procedure for each parameter value between two and ten, in steps of two (see Fig.\ref{fig:SampleImages}-d). For each generated value, if the value is less than the current parameter, we set the corresponding pixel value to 0; if the value is greater than the current parameter, we set the value of the corresponding pixel to 255.

Another effect we consider is the color distortion. To model it, the intensity value of each channel is increased from 20 to 100 in steps of 20 separately, and also the color balance of the image is adjusted with values varying from 5 to 30 in steps of 5. The color balance can be adjusted in camera configurations to render specific colors, particularly neutral colors, correctly (see Fig.\ref{fig:SampleImages}-e and see Fig.\ref{fig:SampleImages}-c).

Finally, we add synthetic occlusion to fixed facial parts. Firstly, to model the sunglasses effect, the dark ellipse shape is fitted on the eye area by using landmark points, which are detected by dlib library \cite{kazemi2014one} (see Fig.\ref{fig:SampleImages}-f left). The rectangle enclosing the eyebrows is fitted onto periocular region (see Fig.\ref{fig:SampleImages}-f right). By using the landmark points, we position rectangles filled with zero values onto nose and mouth area (See Fig.\ref{fig:SampleImages}-g and Fig.\ref{fig:SampleImages}-h). In addition, the scarf effect is modeled by setting zero to all pixels below the nose (see Fig.\ref{fig:SampleImages}-g right).

\section{Results}
The results are reported using two evaluation measures of identification accuracy; Rank-1 and Rank-5 performance. Our obtained baseline results, i.e. results on the LFW closed set identification protocol without having any image degradations on the probe images, are on par with the state-of-the-art face recognition systems when a single representation model is used. For example, the rank-1 face identification result with the single model version of the approach presented in \cite{taigman2015web} is reported as \%72.30, whereas our fine-tuned GoogLeNet model achieves \%77.12 performance on the same protocol. 

All of the used deep network models are found to be very sensitive to image blur. As can be observed in Fig. 3-a and Fig. 3-b, increasing blur level decreases the accuracy dramatically. This decrease in identification accuracy can be explained as the blurring effect removes the edges and smooths color transition on the images. So, it may remove some necessary texture information that the deep CNN model is relying on for discrimination. This indicates that texture information has an important role to learn discriminative features for deep neural networks. 

The deep CNN models are also found to be sensitive to the noise. As can be observed in Fig. 3-a and Fig. 3-b, increasing noise level in the two noise models affects the performance negatively. At the very increased noise level, identification accuracy of the networks becomes less than \%6. However, human eye can still easily recognize subjects at this level as can be seen in Fig.\ref{fig:SampleGaussianEffect}. Another interesting observation is that, the performance of GoogLeNet \cite{szegedy2015going} and AlexNet \cite{krizhevsky2012imagenet} falls faster than VGG-Face \cite{parkhi2015deep} with respect to increasing Gaussian noise. 

Changing the contrast of face images does not affect the performance dramatically. All models show different level of decrease in performance when the contrast is halved or doubled (see Fig. 3-c). As can be seen from Fig. 3-c, the deep CNN models are not robust to brightness distortion. However, they are surprisingly resilient to darkness distortion effect compared to the human eye, which has difficulty to recognize the subjects on darkened images. Increasing color balance causes a slight decrease on the performance (see Fig. 3-f). The performance of GoogLeNet \cite{szegedy2015going} falls faster than VGG-Face \cite{parkhi2015deep} with respect to increasing color balance rate and their performances become equal at distortion level of 30\%.
\begin{figure}[htb]
\IfFileExists{Results2-crop.pdf}{}{\immediate\write18{pdfcrop Results2.pdf}}
\includegraphics[width=\textwidth,keepaspectratio]{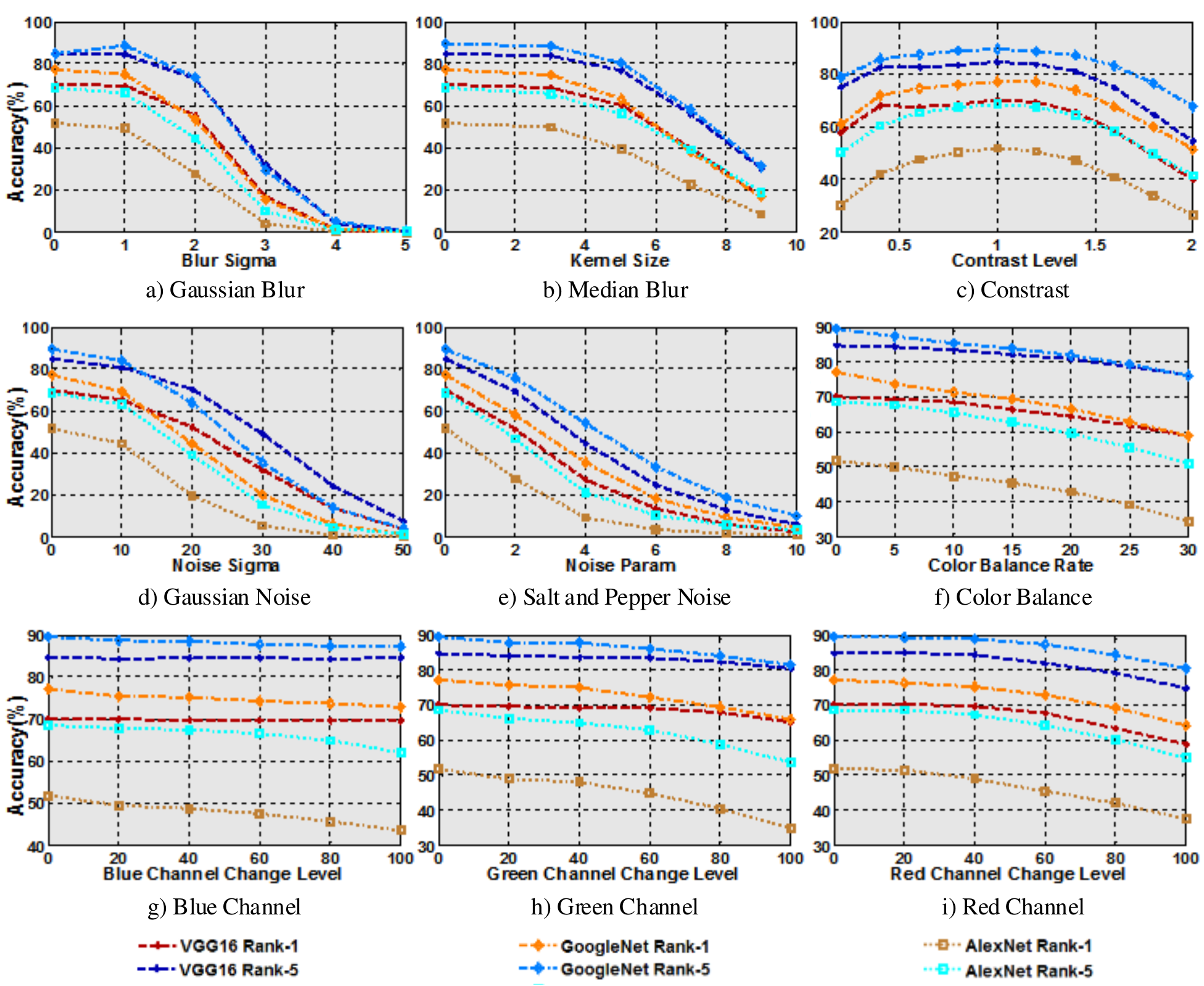}
\caption{Rank-1 and Rank-5 performances of different deep CNN-based face representation under image degradations.}
\label{fig:Results2}
\end{figure}

As can be seen in Fig. 3-g-h-i, the tested networks are robust to increasing intensity of color channels. The performance is affected the most by the change in red and green channel because of skin color. VGG-Face model almost always gives the same performance under increasing intensity of blue channel. AlexNet is the most sensitive model to the same distortion effect.

Tab. 1 shows the results of our experiments with facial part occlusion. The periocular region (see Fig. 2-f right) is found to be the most distinctive area on the face image. This indicates that the networks learn the discriminative features from this area. Eye areas, i.e. sunglasses effect (see Fig. 2-f left), is the second significant facial part. However, on the VGG-Face model, performance is more sensitive to occlusion in nose, than the synthetic sunglasses effect. For all networks, the nose area is found to be more important than the mouth area to learn distinctive features. 

\section{Discussion and Conclusion}
In this paper, we present a comprehensive study to analyze how various kinds of image degradations affect CNN-based face recognition approaches’ performance. The results indicate that image degradations can dramatically lower face recognition accuracy. Especially, blur, noise, and occlusion cause a significant deterioration in the performance. For the occlusion scenario, results show that periocular region is the most important region for successful face recognition. Experimental results also show that deep CNN models are robust to color distortions and change in color balance. According to the results, our fine-tuned GoogLeNet model outperforms other models in the experiments. The margin is almost 10\% between GoogLeNet and VGG-Face in most of the experiments, whereas it can be up to 20\% between GoogLeNet and AlexNet in some of the experiments. 

A practical solution to increase robustness to all of these distortions is to augment the training set with degraded images of each subject. Another solution would be to design models specialized to each kind of degradation, separately. As a future work, we will investigate these approaches and attempt to develop a deep CNN-based face recognition approach that can operate robustly under various image degradations.

\begin{table}[t]
\centering
\begin{tabular}{cccc|cc|cc}
\multicolumn{2}{c}{}&\multicolumn{2}{c}{VGG-Face}&\multicolumn{2}{c}{AlexNet}&\multicolumn{2}{c}{GoogLeNet}\\
\hline
\multicolumn{2}{l}{Occlusion Type} & Rank-1 & Rank-5 & Rank-1 & Rank-5 & Rank-1 & Rank-5\\
\hline
\multicolumn{2}{l}{Sunglasses} & 25.94\% & 41.73\% & 11.85\% & 23.19\% & 31.18\% & 52.72\%\\
\multicolumn{2}{l}{Periocular Region} & 6.33\% & 17.25\% & 2.97\% & 9.39\% & 6.80\% & 21.82\%\\
\multicolumn{2}{l}{Nose} & 20.70\% & 36.68\% & 20.32\% & 34.22\% & 35.59\% & 53.00\%\\
\multicolumn{2}{l}{Mouth} & 37.03\% & 56.77\% & 21.95\% & 41.41\% & 54.66\% & 72.62\%\\
\multicolumn{2}{l}{Scarf Effect} & 41.50\% & 61.95\% & 20.00\% & 38.50\% & 52.52\% & 71.41\%\\
\multicolumn{2}{l}{W/O Occlusion} & 70.13\% & 84.66\% & 51.76\% & 68.53\% & 77.12\% & 89.49\%\\
\hline
\end{tabular}
\caption{\label{tab1}Rank-1 and Rank-5 performances of different deep CNN-based models on the fixed facial part occlusion scenario. The models are most sensitive to occlusion in periocular region.}
\end{table}

\bibliographystyle{lnig}
\bibliography{biosig2016_poster}

\begin{thebibliography}{WHM11}

\bibitem[Ab14]{abaza2014design}
Abaza, Ayman; Harrison, Mary~Ann; Bourlai, Thirimachos; Ross, Arun: Design and
  evaluation of photometric image quality measures for effective face
  recognition.
\newblock IET Biometrics, 3(4):314--324, 2014.

\bibitem[Ba15]{basu2015learning}
Basu, Saikat; Karki, Manohar; Ganguly, Sangram; DiBiano, Robert; Mukhopadhyay,
  Supratik; Nemani, Ramakrishna: Learning sparse feature representations using
  probabilistic quadtrees and deep belief nets.
\newblock In: European Symposium on Artificial Neural Networks, ESANN.
\newblock pp. 367--375, 2015.

\bibitem[Be14]{best2014unconstrained}
Best-Rowden, Lacey; Han, Hu; Otto, Charles; Klare, Brendan~F; Jain, Anil~K:
  Unconstrained face recognition: Identifying a person of interest from a media
  collection.
\newblock IEEE Transactions on Information Forensics and Security,
  9(12):2144--2157, 2014.

\bibitem[CCH14]{chen2014cross}
Chen, Bor-Chun; Chen, Chu-Song; Hsu, Winston~H: Cross-age reference coding for
  age-invariant face recognition and retrieval.
\newblock In: European Conference on Computer Vision.
\newblock Springer, pp. 768--783, 2014.

\bibitem[DK16]{dodge2016understanding}
Dodge, Samuel; Karam, Lina: Understanding How Image Quality Affects Deep Neural
  Networks.
\newblock arXiv preprint arXiv:1604.04004, 2016.

\bibitem[DVS12]{dutta2012impact}
Dutta, Abhishek; Veldhuis, RNJ; Spreeuwers, LJ: The impact of image quality on
  the performance of face recognition.
\newblock 2012.

\bibitem[Hu07]{huang2007labeled}
Huang, Gary~B; Ramesh, Manu; Berg, Tamara; Learned-Miller, Erik: Labeled faces
  in the wild: A database for studying face recognition in unconstrained
  environments.
\newblock Technical report, Technical Report 07-49, University of
  Massachusetts, Amherst, 2007.

\bibitem[Ji14]{jia2014caffe}
Jia, Yangqing; Shelhamer, Evan; Donahue, Jeff; Karayev, Sergey; Long, Jonathan;
  Girshick, Ross; Guadarrama, Sergio; Darrell, Trevor: Caffe: Convolutional
  architecture for fast feature embedding.
\newblock In: Proceedings of the 22nd ACM international conference on
  Multimedia.
\newblock ACM, pp. 675--678, 2014.

\bibitem[JL11]{jain2011handbook}
Jain, Anil~K; Li, Stan~Z: Handbook of face recognition.
\newblock Springer, 2011.

\bibitem[KS14]{kazemi2014one}
Kazemi, Vahid; Sullivan, Josephine: One millisecond face alignment with an
  ensemble of regression trees.
\newblock In: Proceedings of the IEEE Conference on Computer Vision and Pattern
  Recognition.
\newblock pp. 1867--1874, 2014.

\bibitem[KSH12]{krizhevsky2012imagenet}
Krizhevsky, Alex; Sutskever, Ilya; Hinton, Geoffrey~E: Imagenet classification
  with deep convolutional neural networks.
\newblock In: Advances in neural information processing systems.
\newblock pp. 1097--1105, 2012.

\bibitem[NW14]{ng2014data}
Ng, Hong-Wei; Winkler, Stefan: A data-driven approach to cleaning large face
  datasets.
\newblock In: 2014 IEEE International Conference on Image Processing (ICIP).
\newblock IEEE, pp. 343--347, 2014.

\bibitem[PVZ15]{parkhi2015deep}
Parkhi, Omkar~M; Vedaldi, Andrea; Zisserman, Andrew: Deep face recognition.
\newblock In: British Machine Vision Conference.
\newblock volume~1, p.~6, 2015.

\bibitem[SKP15]{schroff2015facenet}
Schroff, Florian; Kalenichenko, Dmitry; Philbin, James: Facenet: A unified
  embedding for face recognition and clustering.
\newblock In: Proceedings of the IEEE Conference on Computer Vision and Pattern
  Recognition.
\newblock pp. 815--823, 2015.

\bibitem[Su15]{sun2015deepid3}
Sun, Yi; Liang, Ding; Wang, Xiaogang; Tang, Xiaoou: Deepid3: Face recognition
  with very deep neural networks.
\newblock arXiv preprint arXiv:1502.00873, 2015.

\bibitem[Sz15]{szegedy2015going}
Szegedy, Christian; Liu, Wei; Jia, Yangqing; Sermanet, Pierre; Reed, Scott;
  Anguelov, Dragomir; Erhan, Dumitru; Vanhoucke, Vincent; Rabinovich, Andrew:
  Going deeper with convolutions.
\newblock In: Proceedings of the IEEE Conference on Computer Vision and Pattern
  Recognition.
\newblock pp. 1--9, 2015.

\bibitem[Ta15]{taigman2015web}
Taigman, Yaniv; Yang, Ming; Ranzato, Marc'Aurelio; Wolf, Lior: Web-scale
  training for face identification.
\newblock In: Proceedings of the IEEE Conference on Computer Vision and Pattern
  Recognition.
\newblock pp. 2746--2754, 2015.

\bibitem[WHM11]{wolf2011face}
Wolf, Lior; Hassner, Tal; Maoz, Itay: Face recognition in unconstrained videos
  with matched background similarity.
\newblock In: Computer Vision and Pattern Recognition (CVPR), 2011 IEEE
  Conference on.
\newblock IEEE, pp. 529--534, 2011.

\bibitem[Yi14]{yi2014learning}
Yi, Dong; Lei, Zhen; Liao, Shengcai; Li, Stan~Z: Learning face representation
  from scratch.
\newblock arXiv preprint arXiv:1411.7923, 2014.

\end{thebibliography}

\end{document}